\documentclass[lettersize,journal]{IEEEtran}
\usepackage{amsmath,amsfonts}
\usepackage{algorithmic}
\usepackage{algorithm}
\usepackage{array}
\usepackage[caption=false,font=normalsize,labelfont=sf,textfont=sf]{subfig}
\usepackage{textcomp}
\usepackage{stfloats}
\usepackage{url}
\usepackage{verbatim}
\usepackage{graphicx}
\usepackage{cite}
\usepackage{booktabs}
\usepackage{amsmath}
\usepackage{multirow}
\usepackage{cleveref}

\begin{document}

\title{GeoRef: Referring Expressions in Geometry via Task Formulation, Synthetic Supervision, and Reinforced MLLM-based Solutions}


\author{
	\IEEEauthorblockN{
        Bing~Liu\IEEEauthorrefmark{1}\textsuperscript{1},
        Wenqiang~Yv\IEEEauthorrefmark{1}\textsuperscript{1},
        Xuzheng~Yang\IEEEauthorrefmark{1}\textsuperscript{1},
        Shichang~Wang\textsuperscript{1},
        Junzhuo~Liu\textsuperscript{1},
        Peng~Wang\IEEEauthorrefmark{2}\textsuperscript{1},
        Guoqing~Wang\textsuperscript{1},~\IEEEmembership{Member,~IEEE,} 
        Yang~Yang\textsuperscript{1},~\IEEEmembership{Senior Member,~IEEE,} 
        and~Heng~Tao~Shen\textsuperscript{1}\textsuperscript{2},~\IEEEmembership{Fellow,~IEEE}}
        
	\IEEEauthorblockA{\textsuperscript{1}University of Electronic Science and Technology of China}
    \IEEEauthorblockA{\textsuperscript{2}Tongji University}
    \thanks{B. Liu, W. Yv and X. Yang contributed equally to this work. P. Wang is the corresponding author.}
} 

\markboth{Journal of \LaTeX\ Class Files,~Vol.~14, No.~8, August~2021}%
{Shell \MakeLowercase{\textit{et al.}}: A Sample Article Using IEEEtran.cls for IEEE Journals}

\IEEEpubid{0000--0000/00\$00.00~\copyright~2021 IEEE}

\maketitle
\begin{abstract}
AI-driven geometric problem solving is a complex vision-language task that requires accurate diagram interpretation, mathematical reasoning, and robust cross-modal grounding. A foundational yet underexplored capability for this task is the ability to identify and interpret geometric elements based on natural language queries. To address this, we introduce the task of Referring Expression Comprehension (REC) for geometric problems, which evaluates whether models can localize points, shapes, and spatial relations in diagrams in response to textual prompts. We present GeoRef, a benchmark dataset constructed from existing geometric problem corpora, featuring diverse, high-quality annotations and queries. Due to the lack of annotated data for this task, we generate a large-scale synthetic training dataset using a structured geometric formal language, enabling broad coverage of geometric concepts and facilitating model adaptation. We explore two fine-tuning approaches: Supervised Fine-Tuning (SFT) and Group Relative Policy Optimization (GRPO). Our results show that GRPO significantly outperforms SFT by better aligning model behavior with task-specific rewards. Furthermore, we propose a verify-and-regenerate mechanism that detects incorrect predictions and re-infers answers using contextual reasoning history, further boosting accuracy. Notably, even state-of-the-art Multimodal Large Language Models (MLLMs) struggle with this task, underscoring the necessity of explicitly evaluating and strengthening geometric grounding as a prerequisite for robust geometric problem solving. Moreover, models trained on GeoRef demonstrate measurable improvements on downstream geometric reasoning tasks, highlighting the broader value of REC as a foundation for multimodal mathematical understanding.
\end{abstract}


\section{Introduction}\label{sec:Introduction}
\begin{figure*}[ht]
  \centering\
   \includegraphics[width=0.9\linewidth]{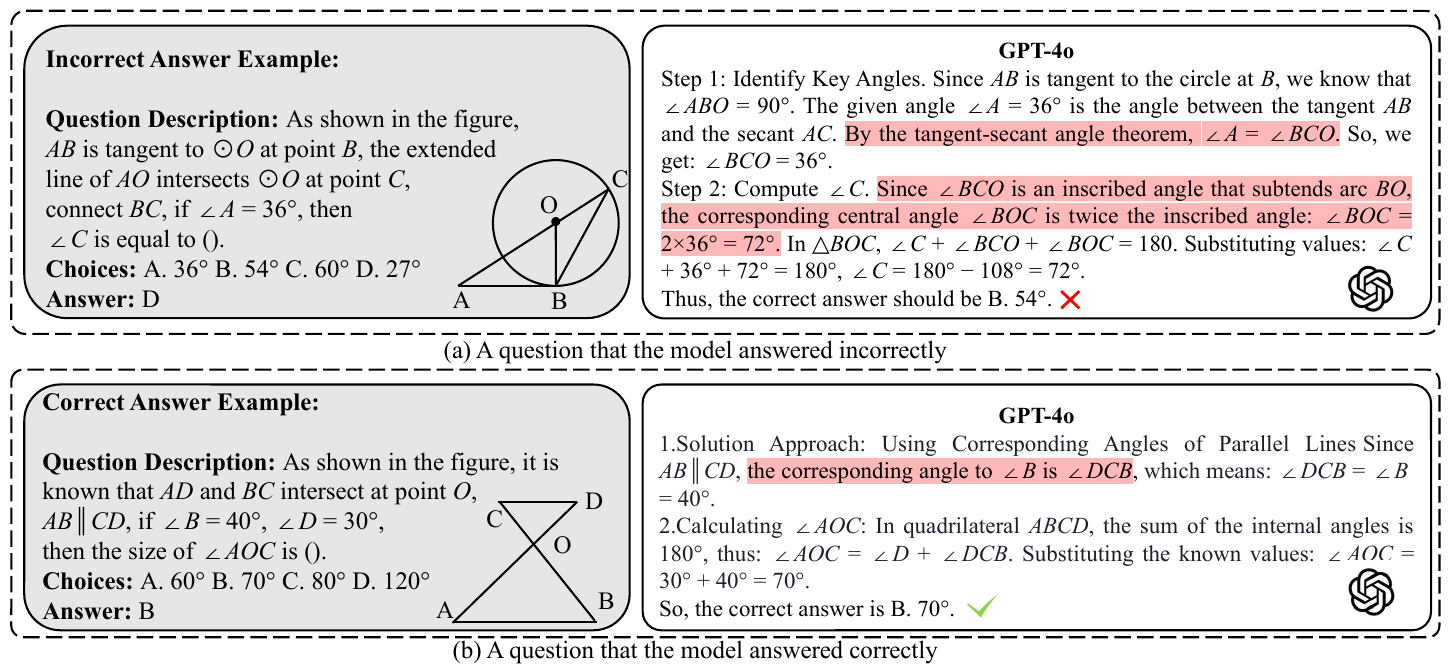}
   \caption{Failure cases where GPT-4o misinterprets geometric elements or relationships, either by (a) providing incorrect answers or (b) giving correct answers. The red text in the figures represents the model's incorrect understanding of the figure. It can be seen that, regardless of whether the model answers the geometric question correctly or not, it may not have understood the geometric figure.}
   \label{fig:motivation}
\end{figure*}

\IEEEPARstart{A}{I} for geometric problem solving presents a unique challenge at the intersection of vision and language, requiring not only logical reasoning but also precise diagram interpretation, spatial understanding, and cross-modal grounding. Unlike purely textual math problems, geometry tasks demand that models identify key elements—such as points, lines, angles, and shapes—understand spatial relationships like intersection and parallelism, and apply relevant geometric theorems in context~\cite{liang2023unimath, zhang2024gold}. While humans naturally follow this process when solving geometric problems, current AI models often bypass it. Existing approaches, especially those based on Multimodal Large Language Models (MLLMs), focus on predicting final answers, offering little insight into whether models genuinely understand the diagram~\cite{gao2023g, shi2024math}, as illustrated in \Cref{fig:motivation}.

To address this gap, we introduce the task of Referring Expression Comprehension (REC)~\cite{mao2016generation,  yu2016modeling} for geometric problems—a novel diagnostic setting designed to assess whether models can correctly identify, interpret, and localize geometric elements based on natural language queries. The REC task isolates the grounding step from full problem solving: It evaluates a model's ability to recognize points (e.g., intersection, midpoint), shapes (e.g., triangles, chords), and spatial relationships (e.g., parallelism, containment) in a diagram.
\IEEEpubidadjcol

To support this task, we present GeoRef, a benchmark dataset designed to rigorously assess MLLMs' geometric comprehension. Built upon the widely used GeoQA corpus~\cite{chen2021geoqa}, GeoRef features high-quality referring expressions annotated across a range of diagrammatic contexts, covering core middle school geometry topics. 
Since no dedicated training dataset exists for this task and manual annotation is highly labor-intensive, we generate a complementary training dataset using a structured geometric formal language. This formal approach ensures that the dataset is scalable, mathematically consistent, and adaptable for training models in geometric comprehension.

To thoroughly evaluate the proposed task and dataset, we benchmark both specialist geometric models and general-purpose MLLMs. To enable effective adaptation, we investigate two fine-tuning strategies: Supervised Fine-Tuning (SFT) and a Group Relative Policy Optimization (GRPO)~\cite{Shao2024DeepSeekMathPT} based reinforced fine-tuning that aligns model predictions with structured task-specific rewards. Additionally, we introduce a verify-and-regenerate mechanism that refines incorrect predictions by conditioning on prior responses and contextual reasoning history, enhancing robustness without explicit ground truth feedback.
Experimental results show that while synthetic supervision significantly improves grounding, MLLMs still face challenges in resolving fine-grained geometric structures and spatial relationships. Crucially, we demonstrate that models trained on the REC task not only improve grounding accuracy but also yield consistent gains on downstream geometric benchmarks—highlighting REC as both a technically meaningful diagnostic tool and a transferable pretraining objective for enhancing mathematical reasoning in multimodal systems.
The contributions of this paper are summarized as follows:

\begin{itemize}
  
\item Task and Dataset: We introduce the Referring Expression Comprehension (REC) task for geometric problem solving and present GeoRef, a high-quality benchmark dataset that systematically evaluates models' ability to identify, interpret, and localize geometric elements and spatial relationships based on natural language queries.

\item Synthetic Data Generation: We design a formal language–driven pipeline to generate large-scale, mathematically consistent synthetic data, enabling effective model training in REC examples.

\item Methodology: We propose a GRPO framework to fine-tune models with structured, reward-driven feedback, and develop a verify-and-regenerate mechanism that enhances model reliability through iterative self-correction based on contextual reasoning history.

\item Empirical Findings and Transferability: Extensive experiments on both specialist and general-purpose MLLMs show that REC remains a challenging task. Importantly, models trained on GeoRef exhibit consistent improvements in downstream geometric reasoning, demonstrating the task’s value as a transferable capability for multimodal mathematical problem solving.
\end{itemize}


\section{Related Work}\label{sec:Related Work}
\subsection{Referring Expression Comprehension}
Referring Expression Comprehension (REC) is a cross-modal task that aims to locate target objects in an image based on natural language descriptions.
Early datasets such as the RefCOCO series~\cite{mao2016generation, yu2016modeling} provided foundational evaluation benchmarks.
Subsequently, datasets based on the GQA scene graph~\cite{hudson2019gqa}, such as Cops-Ref~\cite{chen2020cops} and Ref-Reasoning~\cite{yang2020graph}, introduced compositional reasoning to enrich evaluation criteria.
RefEgo~\cite{kurita2023refego} incorporated negative samples to examine model generalization. FineCops-Ref~\cite{yang2025new} further challenged models with multi-level reasoning complexity.
Existing REC tasks primarily focus on real-world objects, typically using bounding boxes for localization.
However, geometric diagrams differ due to their high abstraction and precision.
Elements such as points, lines, and angles vary in form and are ill-suited for standard bounding box methods.
Understanding geometric diagrams requires reasoning grounded in mathematical definitions and spatial relationships.
To date, REC has yet to systematically explore this complex geometric domain.

\begin{table*}[t]
  \caption{Comparison of geometry benchmarks by task and visual demand.}
  \centering
  \small
  \resizebox{\textwidth}{!}{
  \begin{tabular}{llll}
    \toprule
    \textbf{Dataset} & \textbf{Task Type} & \textbf{Description} & \textbf{Visual Recognition Focus} \\
    \midrule
    PGPS9K~\cite{zhang2023multi}  & Symbolic Geometry Reasoning & Formal symbolic geometry dataset for structured reasoning  & Low \\
    GeoQA~\cite{chen2021geoqa}  & Natural Language Geometry Reasoning & Geometry multiple-choice questions with reasoning steps  & Low \\
    Geometry3K~\cite{lu2021inter} & Numerical Geometry QA & Numerical geometry questions with direct answers  & Low \\
    MathVista~\cite{Lu2023MathVistaEM}  & Visual Mathematical QA & Evaluating MLLMs on visual mathematical reasoning tasks & Moderate \\
    Tangram~\cite{zhang2024tangram} & Counting        & Counting-based visual reasoning with answer counts and rationales        & Moderate \\
    GeoRef(Ours)                    & Referring Expression Comprehension   & Referring expression comprehension over geometric entities and relations & High \\
    \bottomrule
  \end{tabular}
  }
  \label{tab:tab1}
\end{table*}

\subsection{Benchmarks for Geometric Problem Solving}
Geometric problem solving (GPS) remains a core challenge in mathematical AI, requiring the integration of visual comprehension and logical deduction. Current GPS datasets primarily fall into two categories. The first utilizes formal symbolic representations, such as PGPS9K~\cite{zhang2023multi}, supporting structured reasoning but lacking natural language grounding. The second category employs natural language annotations for questions and solutions, exemplified by GeoQA~\cite{chen2021geoqa} and Geometry3K~\cite{lu2021inter}, which are better suited for evaluating language-driven reasoning in multimodal settings.
In addition, recent datasets introduce more focused evaluation angles.
Tangram~\cite{zhang2024tangram} offers a diverse set of 1,080 diagrams and include tasks involving basic geometric object counting.
MathVista~\cite{Lu2023MathVistaEM} and MATH-Vision~\cite{wang2024measuring} are designed to comprehensively evaluate the visual mathematical reasoning capabilities of MLLMs.
Existing datasets lack dedicated evaluation for fine-grained geometric understanding, such as parsing spatial relationships in diagrams. To fill this gap, we propose GeoRef, a benchmark evaluating MLLMs' geometric understanding via referring expressions. It emphasizes geometric grounding, crucial for mathematical vision-language tasks, complementing existing datasets.
A comparison of these datasets is provided in~\Cref{tab:tab1}.

\subsection{Methods for Geometric Problem Solving}
Early approaches to geometric problem solving relied on symbolic logic–based neural methods~\cite{alvin2017synthesis, liang2023unimath}, which first formalized geometric diagrams and then solved them via symbolic reasoning. 
While effective on structured symbolic data, these methods generalize poorly to diverse natural language tasks (e.g., GeoQA) without symbolic annotations.
Recent efforts have turned to MLLMs to better integrate visual and linguistic reasoning. For instance, G-LLaVA~\cite{gao2023g} augmented LLaVA~\cite{liu2023llava} with ChatGPT-generated geometric annotations to form Geo170K. 
These approaches primarily enhance reasoning ability but overlook fundamental geometric elements, leading to inferior performance compared to LLMs. This suggests that current MLLMs fail to effectively leverage their visual modules for geometric structure parsing. We argue that solving geometric problems effectively requires explicit diagram-level grounding, not just end-to-end reasoning.

Following DeepSeek-R1~\cite{guo2025deepseek}, numerous works~\cite{meng2025mm, huang2025vision} have extended reinforcement learning (RL) to multimodal reasoning, integrating self-verification and self-correction to tackle complex tasks via Chain-of-Thought (CoT) reasoning. Other efforts~\cite{zhang2024critic, sun2025mm} focus on constructing verifiers to assess the correctness of answers, but they rely on extensive training and intermediate reasoning annotations, which are costly and challenging to acquire.
In contrast, GeoRef task primarily evaluates the model's grounding and understanding of geometric figures without complex reasoning. 

\section{Task and Dataset}\label{sec:GeoRef}
\subsection{The Task Formulation and GeoRef Dataset}\label{The Task Formulation and GeoRef Dataset}
We introduce a new REC task in the domain of geometric problem solving. Given a geometric image and a textual referring expression (e.g., ``the point where two perpendicular lines intersect''), the objective is to identify the specific geometric element in the image corresponding to the expression. This task evaluates a model's ability to comprehend spatial language, understand geometric semantics, and ground linguistic references in visual geometric content.

To support this task, we present \emph{\textbf{GeoRef}}, a novel dataset designed to assess and enhance the geometric grounding capabilities of MLLMs. GeoRef provides paired data consisting of geometric images, fine-grained descriptions, and referring expressions with corresponding answers. It enables evaluation across multiple categories, from basic element identification to complex spatial relationships.

\textbf{Image Selection.}
GeoRef is built on top of GeoQA ~\cite{chen2021geoqa}, a widely used geometric problem dataset. We manually select images based on the following criteria:
(1) Each image must contain at least two fundamental geometric elements and meaningful spatial relationships (e.g., parallelism, intersection);
(2) The visual structures within the image must be clearly depicted to ensure accurate interpretation;
(3) The visual content must support multi-type question generation—from basic localization (e.g., identifying a point) to relational reasoning (e.g., understanding angle bisectors or perpendicular lines).
This selection process ensures the images comprehensively cover core middle school geometry concepts, as detailed in the supplementary material.

\begin{figure*}[t]

  \centering\
   \includegraphics[width=0.9\linewidth]{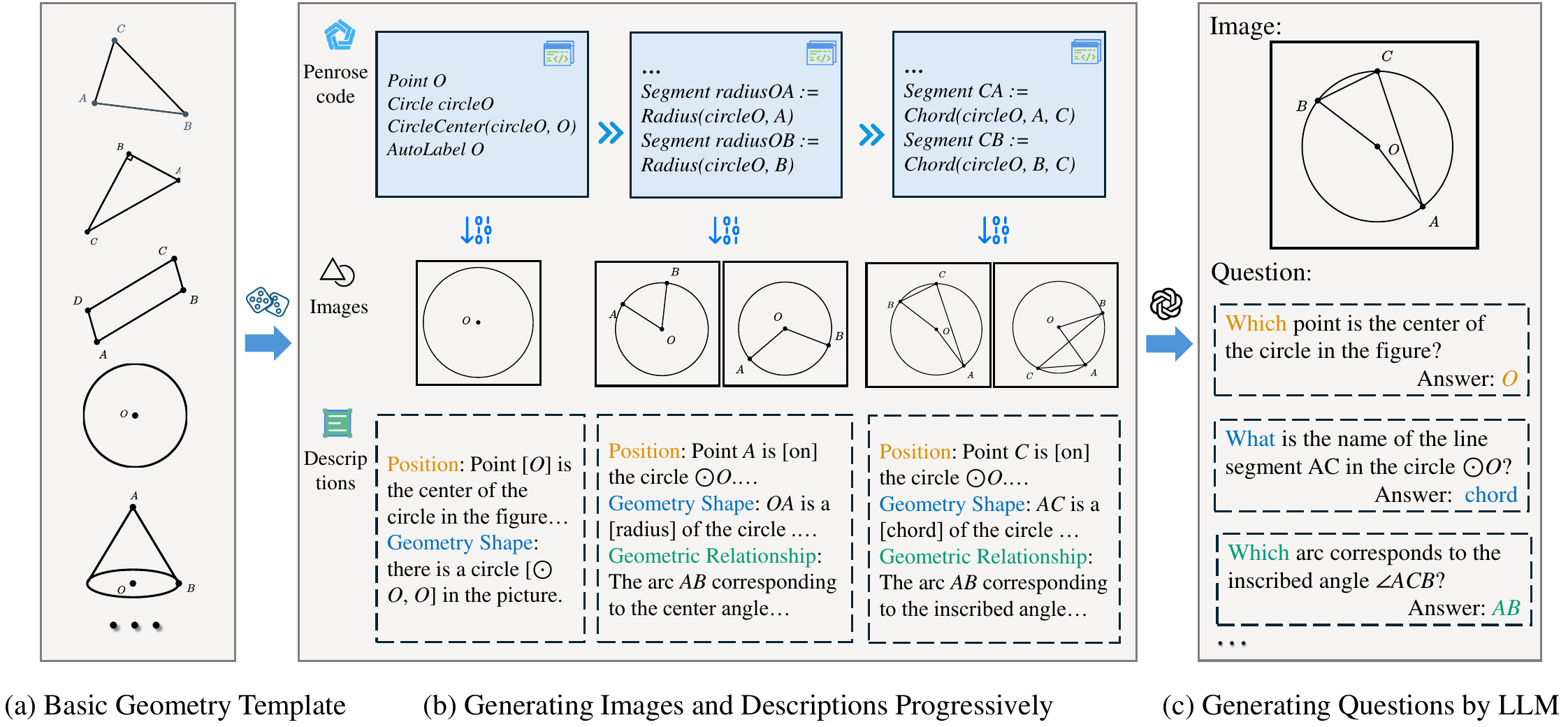}

   \caption{Overview of the dataset synthesis pipeline. First, an initial template is selected from a pool of basic geometric shapes. Next, a geometric element combination related to the template's features is chosen, generating the corresponding geometric diagram and descriptive text progressively through two iterations. Finally, the description is split into individual sentences, and an LLM automatically generates a geometric spatial question for each sentence.}
   \label{fig:pipeline}
   
\end{figure*}

\textbf{Image Description.} To capture the semantic structure of geometric diagrams, we categorize geometric understanding into three types: position, geometric shapes, and geometric relationships.
\emph{\textbf{Position}} refers to the location of key points, which are fundamental units in geometric reasoning. These points fall into two categories: (1) points with intrinsic geometric significance (e.g., the centroid of a triangle), and (2) intersection points formed by geometric elements, such as the intersection of lines \(AB\) and \(CD\). \emph{\textbf{Geometric shapes}} include basic forms such as triangles and circles, as well as contextually significant elements (e.g., a tangent line 
\(AB\) to a circle or a specific triangle in the construction). \emph{\textbf{Geometric relationships}} describe the spatial or logical connections between two or more geometric elements, such as the segments \(AB\) and \(CD\) being parallel or the two angles being equal. 
The selected images are manually annotated according to the three categories, forming the basis of the benchmark dataset.

\textbf{Question and Answer Generation.} We use GLM-4-Flash~\cite{glm2024chatglm} to generate questions based on image descriptions, with prompt templates detailed in the supplementary material. The generation process is guided by a set of carefully defined principles. \textbf{Controlled Question Complexity}:  Most questions are limited to single-step reasoning (e.g., Which angle forms alternate interior angles with angle \(DEA\)?). Few questions require two-step reasoning (e.g., Which angle forms the alternate interior angle with the vertical angle of angle \(DEA\)?). This design choice minimizes confounding factors related to general reasoning ability, allowing the evaluation to focus on geometric understanding. 
\textbf{Open-ended Response}: To discourage random guessing, all evaluation questions adopt an open-ended format. The model must directly generate the geometric element being referred to, rather than selecting from a fixed set of options. 
\textbf{Answer Design and Validation}: Since geometric concepts can be expressed in different but equivalent ways, each question includes a set of acceptable answers. All questions are manually reviewed to ensure correctness, clarity, and consistency, enhancing the reliability and validity of the dataset.

\subsection{Geometric REC Task Synthesis}\label{sec:data_synthesize}

The geometric referring expression comprehension (REC) task aims to localize basic geometric elements within diagrams based on natural language queries. However, existing geometric datasets lack the fine-grained annotations necessary for effective model training, and manual annotation is both time-consuming and difficult to scale. To overcome this limitation, we leverage Penrose~\cite{ye2020penrose}, a declarative diagramming system that enables precise control over the composition and rendering of geometric diagrams from high-level mathematical descriptions. Penrose provides an ideal foundation for synthesizing training data, as it supports flexible generation of diverse diagrams with fine-grained semantic alignment.

Building on this capability, we propose a batch synthesis framework (see~\Cref{fig:pipeline}) that further enhances generation efficiency, visual quality, and geometric diversity. This framework produces a scalable and high-quality resource tailored for training multimodal large language models (MLLMs) in geometric understanding.

\textbf{Geometric Diagram Generation.} 
The dataset synthesis pipeline begins with the selection of a geometric template, such as a triangle or a circle (see~\Cref{fig:pipeline} (a)). For each template, between 4 and 15 sets of formal language representations are incorporated. These formal language sets are manually annotated with detailed descriptions of the geometric elements and their corresponding relationships. The introduction of geometric elements is carried out in two stages. At each stage, it is possible to generate high-quality synthetic geometric images accompanied by comprehensive geometric annotations (see ~\Cref{fig:pipeline} (b)). The resulting dataset systematically encompasses fundamental concepts commonly taught in middle school geometry.

Using this framework, we generate 364 distinct types of synthetic geometric images, each featuring a unique combination of geometric elements and relationships. For each type, we generate one image with common-sense identifiers (e.g., the center of a circle defaults to O) and five with random identifiers, resulting in a total of 2,184 images. Representative examples are provided in the supplementary material. Each image is accompanied by detailed descriptions of its geometric content, categorized into three types: position, geometric shapes, and geometric relationships. 

\textbf{Question Generation.} 
We convert the generated data pairs into question-answer pairs in the same way as~\Cref{The Task Formulation and GeoRef Dataset}. This process results in the training set comprising 2,184 images and 29,815 question-answer pairs.

\begin{figure*}[t]

  \centering\
   \includegraphics[width=0.75\linewidth]{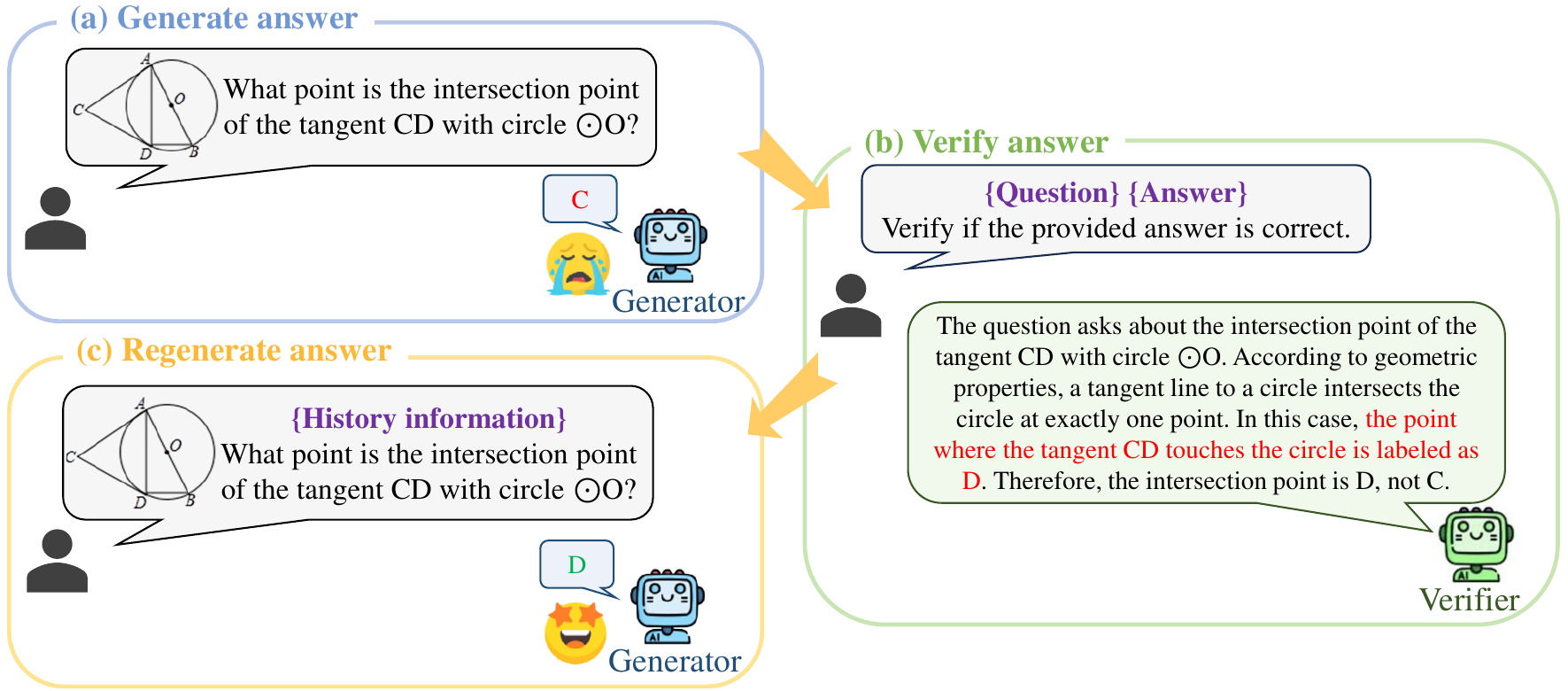}

   \caption{Illustration of the idea of the proposed Verify-and-Regenerate mechanism. Given an image-question pair, \textbf{Generator} first generates an initial answer. This image-question-answer triplet is then passed to \textbf{Verifier}, which evaluates the correctness of the generated answer and provides reasoning feedback. Finally, using the dialogue from the previous two stages as contextual history, \textbf{Generator} refines and regenerates the answer to produce the final output.}
   \label{fig:method}
\end{figure*}

\section{Method}\label{sec:Method}
In this section, we introduce two fine-tuning strategies for improving performance on the geometric REC task: SFT and GRPO-based Reinforced Fine-Tuning (RFT). Specifically, we design a rule-based accuracy reward function tailored for GRPO to align model predictions with geometric correctness. In addition, we propose a verify-and-regenerate mechanism, which identifies incorrect predictions and re-generates answers by leveraging contextual reasoning history, thereby further enhancing overall accuracy.

\subsection{Training Methods}
\textbf{Supervised Fine-Tuning.} SFT leverages a labeled dataset of question–answer pairs to adapt the base model toward outputs that closely match ground-truth responses. This approach relies on direct supervision and learns by minimizing the difference between generated and target answers.

\textbf{Reinforced Fine-Tuning.} RFT learns from reward signals instead of explicit labels. Since the model is not given the correct answer during training, RFT is less susceptible to overfitting or memorization. In this work, we adopt GRPO~\cite{Shao2024DeepSeekMathPT}, a lightweight and effective reinforcement learning algorithm that has shown strong empirical performance and requires minimal hyperparameter tuning. GRPO compares the relative quality of multiple responses to the same prompt, allowing the model to learn preferences in a sample-efficient manner. 

We design a binary rule-based accuracy reward function to evaluate correctness of model outputs. The reward function takes a question–output pair and assigns a reward of 1 if the prediction belongs to the ground-truth answer set, and 0 otherwise.


\subsection{Verify-and-Regenerate Mechanism}
\Cref{fig:method} presents an overview of the proposed verify-and-regenerate mechanism. Given an input consisting of an image-question pair, the process begins with the generator, which produces an initial answer based on the visual and textual inputs. This forms an image-question-answer triplet, which is subsequently fed into the verifier. 

The verifier checks the answer's validity by providing reasoning and a binary judgment, identifying any inaccuracies or ambiguities. Finally, the generator leverages the full dialogue history—including the original image–question pair, the initial answer, and the verifier’s feedback—as contextual input to regenerate a new answer. This leads to a revised and more accurate final response, benefiting from explicit verification and guided refinement.

In our implementation, we utilize the generator trained with GRPO, which enhances answer generation, while the verifier is instantiated as the base model without additional training. This task-specific role division fosters complementary collaboration: the generator proposes potential answers, and the verifier refines them through validation. Together, they form a feedback loop that significantly improves overall answer quality and robustness.

\begin{table*}[t]

 \centering
   \caption{Accuracy (\%) of different models on the GeoRef dataset. The comparison includes off-the-shelf base models, models fine-tuned on the proposed synthesized data, and human performance. The bolded values indicate the best results.}   
 \resizebox{0.75\textwidth}{!}{
 \normalsize
 \begin{tabular}{lcccc}
 \toprule
 \multirow{2}{*}{\textbf{Model}} & \multirow{2}{*}{\textbf{Position}} & \textbf{Geometry} & \textbf{Geometric} & \textbf{Overall} \\
 & & \textbf{Shape} & \textbf{Relationship} & \textbf{Accuracy} \\
 \midrule
  \textbf{Base Models} &  &  &  &  \\
  LLaVA-1.5-7B~\cite{liu2023llava} & 14.72 & 27.08 & 7.88 & 16.22 \\
 Math-LLaVA-13B~\cite{shi2024math}& 18.44 & 29.12 & 13.64 & 20.22 \\
 G-LLaVA-13B~\cite{gao2023g} & 10.34 & 37.81 & 15.09 & 22.39 \\
 LLaVA-v1.6-Mistral-7B~\cite{liu2024llavanext} & 21.35 & 32.19 & 22.42 & 25.76 \\
 Llama-3.2-11B-Vision-Instruct~\cite{chi2024llama} & 29.58 & 38.25 & 33.54 & 34.45 \\
 MiniCPM-V-2.6~\cite{yao2024minicpm} & 43.37 & 49.71 & 27.78 & 38.85 \\
 Qwen2.5-VL-7B~\cite{qwen2.5-VL} & 46.95 & 51.46 & 55.15 & 52.17 \\
 InternVL2.5-8B~\cite{chen2024expanding} & 47.88 & 67.08 & 46.91 & 54.42 \\
 GPT-4o~\cite{GPT-4o} & 55.10 & 62.48 & 49.18 & 55.26 \\
 Qwen2.5-VL-72B~\cite{qwen2.5-VL} & \textbf{70.29} & 57.93 & 59.69 & 61.17 \\
 \midrule
 \textbf{Models Trained on Synthesized Data} &  &  &  &  \\
 Qwen2.5-VL-7B-SFT & 55.17 & 64.89 & 57.87 & 59.88 \\
 Qwen2.5-VL-7B-GRPO & 56.10 & 68.18 & \textbf{62.89} & 63.45 \\
 Qwen2.5-VL-7B-GRPO-verify & 62.47 & \textbf{71.31} & 62.35 & \textbf{65.62} \\
 \midrule
 Human & 96.69 & 96.57 & 93.52 & 95.26 \\
 \bottomrule
 \end{tabular}
 }

  \label{tab:evaluation_results}
\end{table*}
\section{Experiments}\label{sec:Experiments}
In this section, we present our experimental results.~\Cref{sec:Dataset} details dataset statistics. In~\Cref{sec:benckmark}, we benchmark representative models on the GeoRef dataset, highlighting core challenges and characteristics of the REC task. Meanwhile, we evaluate the effectiveness of the proposed verify-and-regenerate mechanism. Finally,~\Cref{sec:Ablation} presents ablation studies that further validate the impact of our methodological components and the generalization of GeoRef on downstream geometric problems. 

\subsection{Dataset}\label{sec:Dataset}
\subsubsection{\textbf{Test Dataset GeoRef}}
The GeoRef test set comprises 3,776 questions, categorized into three types. Each instance includes a geometric diagram, a natural language question, and a predefined answer pattern. To ensure the reliability of evaluation, all test instances are manually verified for accuracy in both visual content and linguistic annotations. During testing, the model receives a question and its corresponding diagram as input and is expected to produce a direct answer grounded in the visual content. Model predictions are evaluated by applying regular expressions to extract and match key answer components. Final performance is reported using accuracy.

\begin{table}[t] 
  \centering
  \caption{Statistics of the proposed datasets.}
  \small
  \resizebox{0.4\textwidth}{!}{
  \begin{tabular}{@{}lcc@{}}
    \toprule
             & GeoRef & Training Set \\
    \midrule
    Position & 755 & 7433 \\
    Geometry Shape & 1370 & 8395 \\
    Geometric Relationship & 1651 & 14076 \\
    \midrule
    Total Questions & 3776 & 29815 \\
    Images & 392 & 2184 \\
    \bottomrule
    
  \end{tabular}
  }
  
  \label{tab:statistics}
\end{table}

\subsubsection{\textbf{Synthesized Training Dataset}}
We use the synthesized data introduced in~\Cref{sec:data_synthesize} as the training dataset. Each entry in the training dataset consists of a question, an answer and an image. ~\Cref{tab:statistics} presents the statistics of our datasets.

\subsection{Benchmarking the GeoRef Dataset}\label{sec:benckmark}
We evaluate the geometric spatial understanding of various MLLMs on the GeoRef dataset, including both geometry-specialized and general-purpose models. Implementation details and training setting are provided in the supplementary material. ~\Cref{tab:evaluation_results} presents the evaluation results. To provide a human reference, we also conducted a human evaluation with graduate students from science and engineering backgrounds. Interestingly, even humans didn’t perform perfectly—not due to annotation errors, but occasional misinterpretation of geometric concepts—highlighting the task's difficulty.

\subsubsection{\textbf{Limitations of MLLMs on the GeoRef Benchmark}}
Among the evaluated models, the large-scale open-source Qwen2.5-VL-72B achieved the highest accuracy at 61.17\%, followed by the closed-source GPT-4o at 55.26\%. Among open-source general-purpose models (7B–13B), InternVL2.5-8B and Qwen2.5-VL-7B showed strong performance, approaching GPT-4o. In contrast, MiniCPM-V-2.6 and Llama-3.2-11B performed considerably worse. Surprisingly, geometry-specialized models such as G-LLaVA-13B and Math-LLaVA-13B underperformed even the weakest general-purpose models, suggesting that their training focused heavily on symbolic mathematical reasoning while neglecting geometric grounding and visual localization. Similarly, earlier models like LLaVA-v1.5-7B and LLaVA-v1.6-Mistral-7B showed clear limitations on this task.

Despite the top-performing model achieving over 60\% accuracy, the results reveal a substantial gap in MLLMs’ geometric spatial understanding. Position recognition exhibited the largest performance variance (14.72\%–70.29\%), indicating inconsistent grounding of visual elements. Geometric shape recognition was the most successful subtask, while geometric relation reasoning (e.g., identifying parallel or perpendicular relationships) proved most challenging. This contrast underscores the cognitive hierarchy within geometric spatial tasks—basic object identification is achievable, whereas reasoning over spatial relations poses a significant challenge for current models. Additional qualitative visualization examples are provided in the supplementary material.

\subsubsection{\textbf{GRPO Significantly Outperforms SFT}}
As shown in~\Cref{tab:evaluation_results}, both SFT and GRPO fine-tuning substantially improve Qwen2.5-VL-7B’s performance on the GeoRef task. However, GRPO consistently outperforms SFT, achieving a +3.57\% gain under the same training setting. Remarkably, GRPO even surpasses Qwen2.5-VL-72B, demonstrating its superior sample efficiency. The performance boost varies across sub-tasks: GRPO improves position recognition by nearly 10\% and geometric shape recognition by 15\%, highlighting the model’s ability to parse spatial structures. These results suggest that existing MLLMs lack sufficient grounding supervision in their pretraining, and that the GeoRef task—combined with a reward-driven adaptation strategy like GRPO—can meaningfully bridge this gap in geometric visual understanding.

\subsubsection{\textbf{Verify-and-Regenerate Mechanism Further Boosts Accuracy upon GRPO}}
Building on the Qwen2.5-VL-7B model fine-tuned with GRPO, we apply the proposed verify-and-regenerate mechanism, resulting in the enhanced  Qwen2.5-VL-7B-GRPO-verify variant. As shown in~\Cref{tab:evaluation_results}, this mechanism further enhances accuracy, establishing a new state-of-the-art among all evaluated models.
Notable improvements are seen in tasks involving Position and Geometry Shape, while performance on Geometric Relationship remains stable. This suggests that the mechanism's particular effectiveness in tasks requiring localized visual recognition and element-level grounding. These results highlight the mechanism’s effectiveness in tasks that demand a finer-grained understanding of visual elements and spatial relationships.
\begin{table*}[th]
 \centering
  \caption{Accuracy (\%) of different MLLMs fine-tuned on our synthesized training set and evaluated on the GeoRef dataset.}

 \resizebox{0.8\textwidth}{!}{
 \normalsize	
 \begin{tabular}{lcccc}
 \toprule
 {\textbf{Model}} & {\textbf{Position}} & {\textbf{Geometry Shape}} & {\textbf{Geometric Relationship}} & {\textbf{Overall Accuracy}} \\
 \midrule
  MiniCPM-V-2.6 & 43.37\ & 49.71\ & 27.78\ & 38.85\ \\
 MiniCPM-V-2.6-SFT & {\textbf{53.45 ($\uparrow$9.75)}} & {\textbf{57.01 ($\uparrow$7.3)}} & {\textbf{49.33 ($\uparrow$21.55)}} & {\textbf{52.94 ($\uparrow$14.09)}} \\

 \midrule

 InternVL2.5-8B & 47.88\ & 67.08\ & 46.91\ & 54.42\ \\
 InternVL2.5-8B-SFT & {\textbf{49.20 ($\uparrow$1.32)}} & {\textbf{71.89 ($\uparrow$4.81)}} & {\textbf{56.84 ($\uparrow$9.93)}} & {\textbf{60.81 ($\uparrow$6.39)}} \\
 \midrule   
 Qwen2.5-VL-7B & 46.95\ & 51.46\ & 55.15\ & 52.17\ \\
 Qwen2.5-VL-7B-SFT & {\textbf{55.17 ($\uparrow$8.22)}} & {\textbf{64.89 ($\uparrow$13.43)}} & {\textbf{57.87 ($\uparrow$2.75)}} & {\textbf{59.88 ($\uparrow$7.71)}} \\
 \bottomrule
 \end{tabular}
 }

 \label{tab:accuracy_of_different_MLLMs}
\end{table*}

\begin{table*}[ht]
 \centering
\caption{Ablation study on the verify-and-regenerate mechanism.
\textit{Baseline} refers to the Qwen2.5-VL-7B model fine-tuned with GRPO.
\textit{Generation from Verifier} denotes a variant where the verifier not only verifies but also directly generates the final answer.
\textit{Verify-and-Regenerate} represents our proposed method, where the verifier provides feedback and the generator refines the answer based on reasoning history.}
 
  \resizebox{0.85\textwidth}{!}{
 \normalsize	
 \begin{tabular}{lcccc}
 \toprule
 \textbf{Method} & \textbf{Position} & \textbf{Geometry Shape} & \textbf{Geometric Relationship} & \textbf{Overall Accuracy} \\
 \midrule
 Baseline & 56.10 & 68.18 & \textbf{62.89} & 63.45 \\
 Generation from Verifier & 58.75 & 67.88 & 59.50 & 62.39 \\
 Verify-and-Regenerate & \textbf{62.47} & \textbf{71.31} & 62.35 & \textbf{65.62} \\
 \bottomrule
 \end{tabular}
 }
 
\label{tab:ablation}
\end{table*}

\begin{table*}[ht]
\centering
\caption{Accuracy (\%) comparison of different methods on downstream geometric problems. For GeoQA, the evaluation follows the settings in ~\cite{chen2025r1v}. For MathVista and MATH-Vision, we follow the same evaluation protocol suggested in MM-Eureka~\cite{meng2025mm}. The bolded values indicate the best results.}
 \resizebox{0.85\textwidth}{!}{
\begin{tabular}{lccccccc} 
\toprule
\multirow{3}{*}{\textbf{Method}} & \multirow{3}{*}{\textbf{GeoQA}}              & \multicolumn{3}{c}{\textbf{ MathVista }}                                                                       & \multicolumn{3}{c}{\textbf{MATH-Vision}}                                                                       \\ 
\cmidrule(lr){3-5}\cmidrule(lr){6-8}
                                 &                                              & \textbf{\textbf{Geometry}} & \textbf{\textbf{Geometry~}}                         & \textbf{\textbf{Algebraic}} & \multicolumn{3}{c}{\textbf{\textbf{metric geometry}\textbf{~-}}}                                                                             \\
                                 &                                              & \textbf{Reasoning}         & \textbf{\textbf{\textbf{\textbf{Problem~}}}Solving} & \textbf{Reasoning}          & \textbf{angle} & \textbf{area} & \textbf{~length}  \\ 
\midrule
Qwen2.5-VL-7B                    & 46.81          & 58.58                       & 58.65                             & 59.07                        & 31.79                                                          & 24.40                                                          & 27.83                                                            \\
Qwen2.5-VL-7B-SFT      & 47.61          & 60.25                       & 61.05                             & 60.14                        & 27.74                                                          & 26.00                                                          & 28.73                                                            \\
Qwen2.5-VL-7B-GRPO      & \textbf{53.84} & \textbf{63.60}              & \textbf{63.94}                    & \textbf{63.34}               & \textbf{33.52}                                                 & \textbf{27.80}                                                 & \textbf{30.95}                                                   \\
\bottomrule
\end{tabular}
}

\label{tab:mathvista_performance}
\end{table*}

\begin{table}[h]
  \centering
    \caption{Accuracy (\%) of models evaluated on the \(Common\) and \(Random\) datasets. Models fine-tuned on the Cmn, Rand, and Hybrid annotated training sets.}
   \resizebox{0.4\textwidth}{!}{
  \label{tab:bias}
  \normalsize
  \begin{tabular}{lcc}
    \toprule
    \textbf{Model} & \textbf{\(Common\)} & \textbf{\(Random\)} \\
    \midrule
    Qwen2.5-VL-72B & 53.11 & 41.33 \\
    \midrule
    Qwen2.5-VL-7B & 42.89 & 30.89 \\
    Qwen2.5-VL-7B-Cmn & 66.22 & 38.22 \\
    Qwen2.5-VL-7B-Rand & 57.78 & 40.22 \\
    Qwen2.5-VL-7B-Hybrid & 64.00 & 39.78 \\
    \bottomrule
  \end{tabular}}
\end{table}

\subsection{Ablation Study}
\label{sec:Ablation}
\subsubsection{\textbf{Effectiveness of Synthesized Data}} \label{Effectiveness}
Fine-tuning on the synthesized dataset yields consistent performance gains across all models (\Cref{tab:accuracy_of_different_MLLMs}). MiniCPM-V-2.6 benefits most, with notable improvements in relational reasoning, while Qwen2.5-VL-7B exhibits balanced gains across tasks, highlighting strong adaptability. InternVL2.5-8B shows more moderate but targeted improvements, particularly in relational reasoning.

Overall, these results indicate that synthetic supervision is an effective and scalable strategy for strengthening spatial understanding. They further suggest that architectural design and pretraining biases influence how models absorb grounding signals, and that multi-task fine-tuning enhances cross-task transfer while mitigating the limitations of single-task training.
\subsubsection{\textbf{Bias due to Diagram Annotation}} \label{Bias}
We examine the effect of annotation conventions in geometric diagrams, where certain letters (e.g., using \(O\) for a circle center) are more commonly adopted than arbitrary assignments. To isolate this factor, we constructed two evaluation sets of equal size and difficulty: $Common$ (standard labels) and $Random$ (arbitrary labels). We then fine-tuned Qwen2.5-VL-7B on three training sets of identical size (4,905 problems each): Cmn (standard labels), Rand (arbitrary labels), and Hybrid (a 1:1 mix).  Results show that fine-tuning with a single annotation scheme amplifies bias, improving performance on the matched dataset but reducing generalization to the other. In contrast, hybrid supervision yields more balanced outcomes, effectively mitigating bias and enhancing robustness in geometric reasoning.

\subsubsection{\textbf{Verify-and-Regenerate vs. Generation from Verifier}}
As shown in~\Cref{tab:ablation}, we compare three setups to analyze the components of the verify-and-regenerate mechanism. In the \textit{Generation from Verifier} setting, the generator produces an initial answer, and the verifier is prompted to directly provide the corrected answer using \textless new\_answer\textgreater \ tags. In contrast, our verify-and-regenerate method (i.e., Qwen2.5-VL-7B-GRPO-verify) incorporates the verifier’s feedback as contextual input and re-invokes the generator to produce a refined response. This design leverages the full interaction history between the generator and verifier, yielding better alignment and more accurate final outputs. 

\subsubsection{\textbf{Generalization on Downstream Geometric Problems}}
~\Cref{tab:mathvista_performance} presents results on downstream geometric problems, including GeoQA~\cite{chen2021geoqa}, as well as the plane geometry subsets from MathVista\cite{Lu2023MathVistaEM} and MATH-Vision~\cite{wang2024measuring}.
Nearly all fine-tuned variants of Qwen2.5-VL-7B outperform the untuned baseline, demonstrating the transferability of REC training to downstream mathematical reasoning tasks. This highlights the broader value of the proposed REC task as a foundation for multimodal mathematical understanding. Regarding fine-tuning strategy, GRPO outperforms SFT consistently under identical settings, achieving up to +3.64\% absolute gain. This suggests that GRPO better aligns model behavior with the structural demands of geometric reasoning.
\section{Conclusion }
\label{Conclusion}
We present a new task and benchmark for Referring Expression Comprehension (REC) in geometric problems, along with GeoRef, a semantically diverse dataset that covers point identification, shape recognition, and spatial relationships. To address the lack of training data, we synthesized a large-scale dataset using a formal geometric language and proposed effective fine-tuning strategies—both supervised and GRPO-based reinforcement learning—augmented by a novel verify-and-regenerate mechanism. Our methods significantly improve grounding accuracy and robustness, and models trained on geometric REC dataset exhibit consistent gains on downstream geometric problem-solving tasks, confirming the task’s generalizability and practical value.
\bibliographystyle{IEEEtran}
\bibliography{TPAMI}
\clearpage

\appendices

\section{Additional Details of GeoRef}

\begin{figure*}[ht]
  \centering\
   \includegraphics[width=0.8\linewidth]{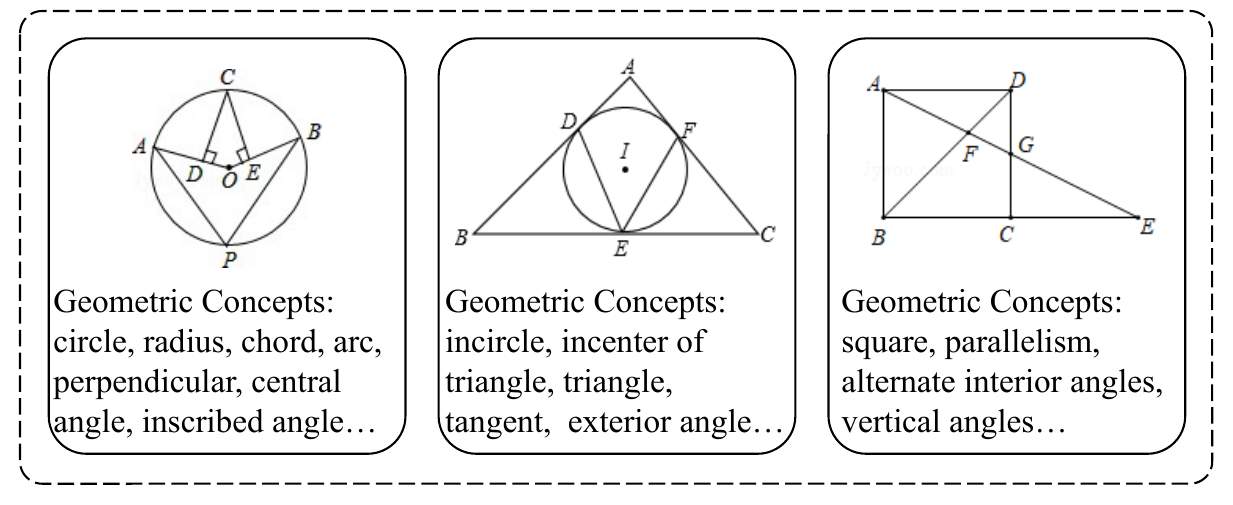}

   \caption{Every image is composed of multiple geometric concepts.}
   \label{fig:fig5}
\end{figure*}

\begin{figure}[h]
  \centering
  \includegraphics[width=1\linewidth]{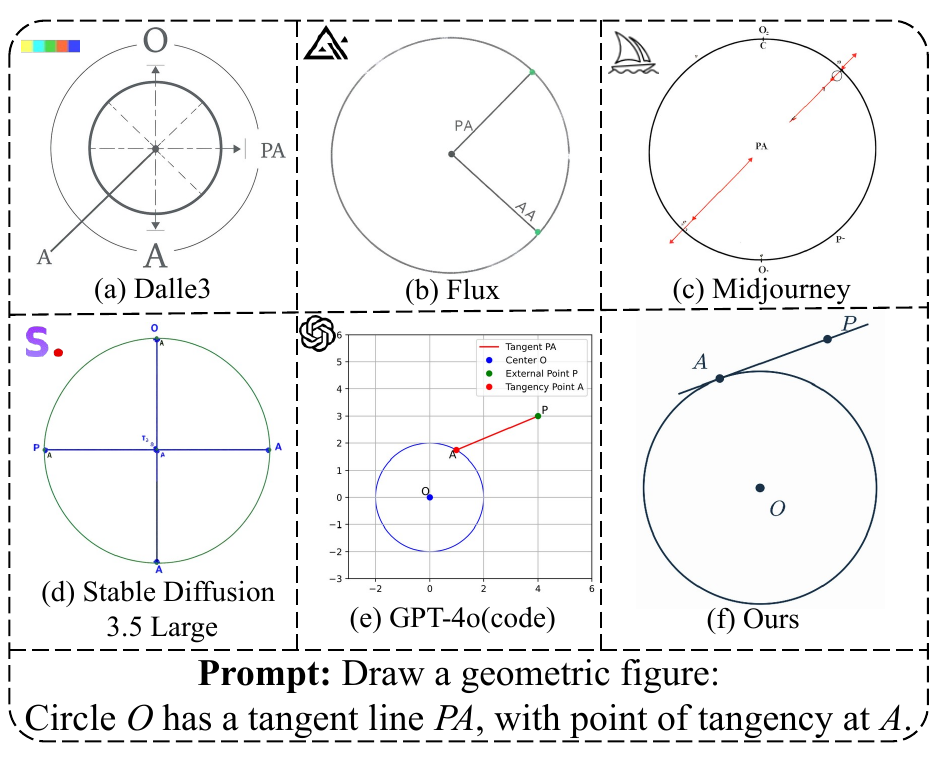}
  \caption{Examples of generated geometric diagrams. (a)–(d) show results from text-to-image models; (e) is generated by GPT-4o~\cite{GPT-4o} via code, and (f) by our method.}
  \label{fig:fig3}
\end{figure}

\begin{figure}[h]
\centering
\includegraphics[width= 0.8 \columnwidth]{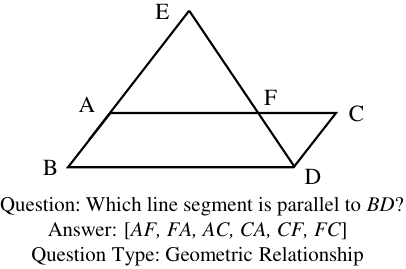}
\caption{An example from GeoRef. The answer to a geometry question may be expressed in multiple equivalent forms, forming an answer set. Since all expressions refer to the same geometric element, any response within the set is considered correct.}
\label{fig:fig2}
\end{figure}

\begin{figure}[h]
  \centering\
   \includegraphics[width=1\linewidth]{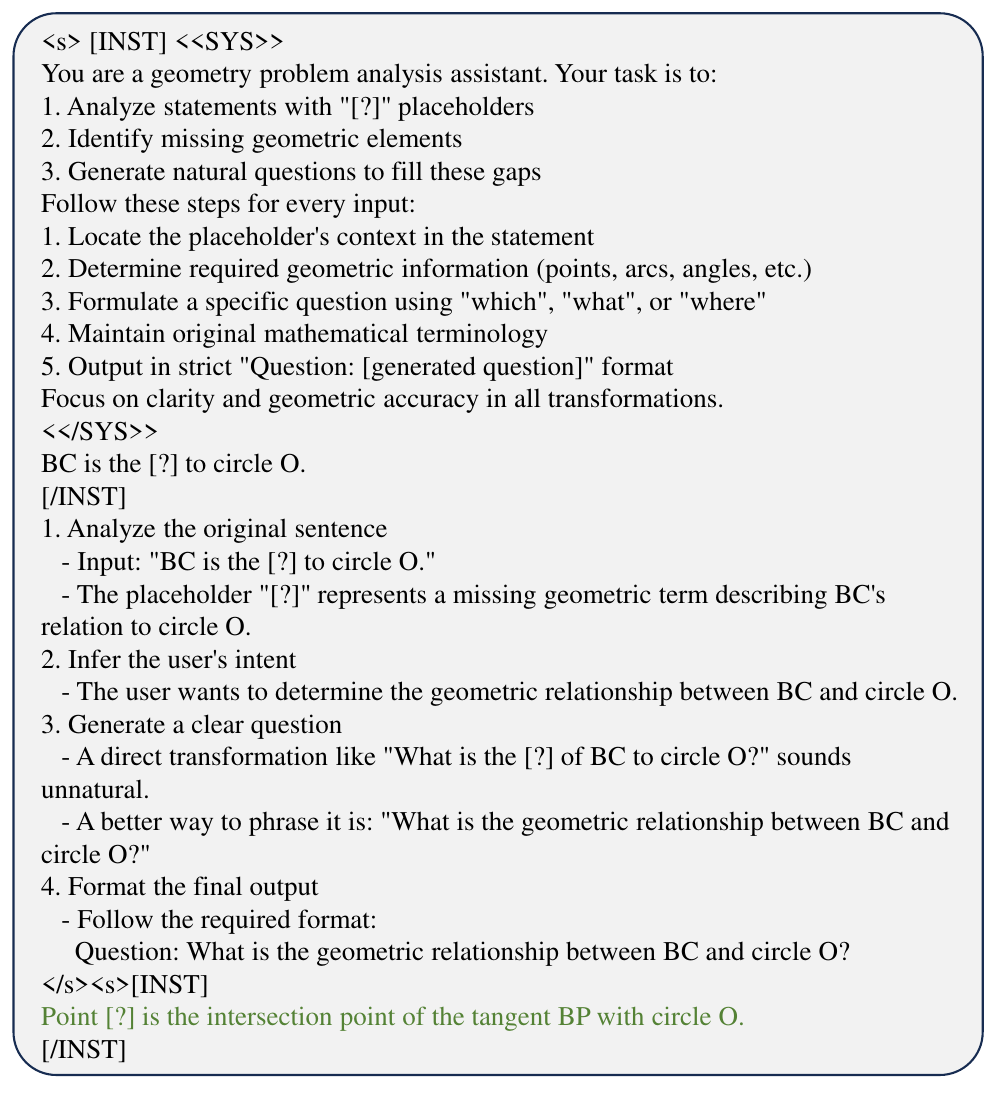}
   \caption{Prompt for question generation.}
   \label{fig:prompt}
\end{figure}




\subsection{Comprehensive Geometric Concepts Coverage in the Proposed Datasets\label{concepts}}

The images in both the proposed GeoRef dataset and the synthesized dataset encompass a comprehensive range of geometric shapes (e.g., triangles, circles, squares), elements (e.g., radius, central angle, tangent lines), and relationships (e.g., parallel, perpendicular bisector, alternate interior angles) in middle school plane geometry, covering a total of 42 distinct geometric concepts. Each image integrates compositional geometric concepts, as illustrated in~\Cref{fig:fig5}.

\subsection{Limitations of Diffusion Models in Geometric Generation}\label{sec:diffusion}

Diffusion models (e.g., Stable Diffusion~\cite{rombach2022high}, DALL-E~\cite{betker2023improving}, Midjourney~\cite{midjourney}, Flux~\cite{flux2024}) often fail to produce geometrically valid outputs, even when provided with precise prompts (e.g., ``Draw a circle 
\(O\) with a tangent line 
\(PA\), where
\(A\) is the point of tangency''), as illustrated in~\Cref{fig:fig3}.

\subsection{Question Generation}\label{sec:prompt}

An example of a generated question and its answ is shown in~\Cref{fig:fig2}. When describing the image, we enclose key geometric elements in square brackets, such as ``BC is the [tangent] to circle O.'' During question generation, we first replace the key elements in each sentence of the description with ``?'', and then use the prompt shown in~\Cref{fig:prompt} to guide the model in generating questions.

\subsection{Examples of Synthetic Geometric Images}\label{sec:Examples}
As shown in~\Cref{fig:synthetic}, although these images are all constructed based on the same formal language—which consistently employs circles, diameters, and inscribed angles as fundamental geometric primitives—they exhibit significant diversity in their overall visual forms. This characteristic not only ensures a high degree of visual variability within the dataset, but also provides abundant and representative training resources for geometric understanding tasks, thereby enhancing the generalization ability and robustness of learning models.

\IEEEpubidadjcol

\section{Experimental Setup}\label{Experimental Setup}

\textbf{Hardware Information.}
All experiments are run on a machine with an Intel(R) Xeon(R) Gold 6348 CPU with a 512G
memory and four 80G NVIDIA RTX A800 GPUs.

\textbf{Models for Benchmarking.}
We evaluated a variety of MLLMs, which can be categorized as follows:


\begin{itemize}
    \item Open-source general-purpose MLLMs: LLaVA-v1.6-Mistral-7B~\cite{liu2024llavanext}, LLaVA-v1.5-7B~\cite{liu2023llava}, Llama-3.2-11B-Vision-Instruct~\cite{chi2024llama}, MiniCPM-V-2.6~\cite{yao2024minicpm}, InternVL-2.5-8B~\cite{chen2024expanding}, Qwen2.5-VL-7B, and Qwen2.5-VL-72B~\cite{qwen2.5-VL}.
    
    \item Closed-source general-purpose MLLMs: GPT-4o~\cite{GPT-4o}.
    
    \item Open-source expert MLLMs: Math-LLaVA-13B~\cite{shi2024math} and G-LLaVA-13B~\cite{gao2023g}.
\end{itemize}

\begin{figure}[h]
  \centering\
   \includegraphics[width=0.8\linewidth]{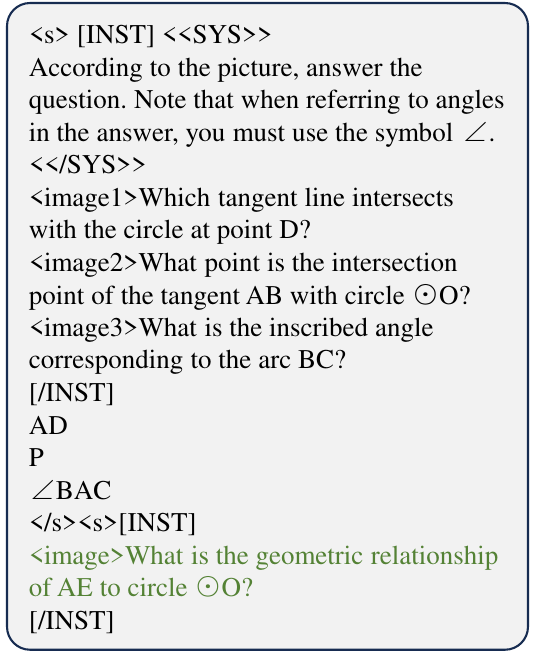}
   \caption{Prompt for evaluation.}
   \label{fig:prompt-qa}
\end{figure}

\begin{figure*}[h]
  \centering\
   \includegraphics[width=0.6\linewidth]{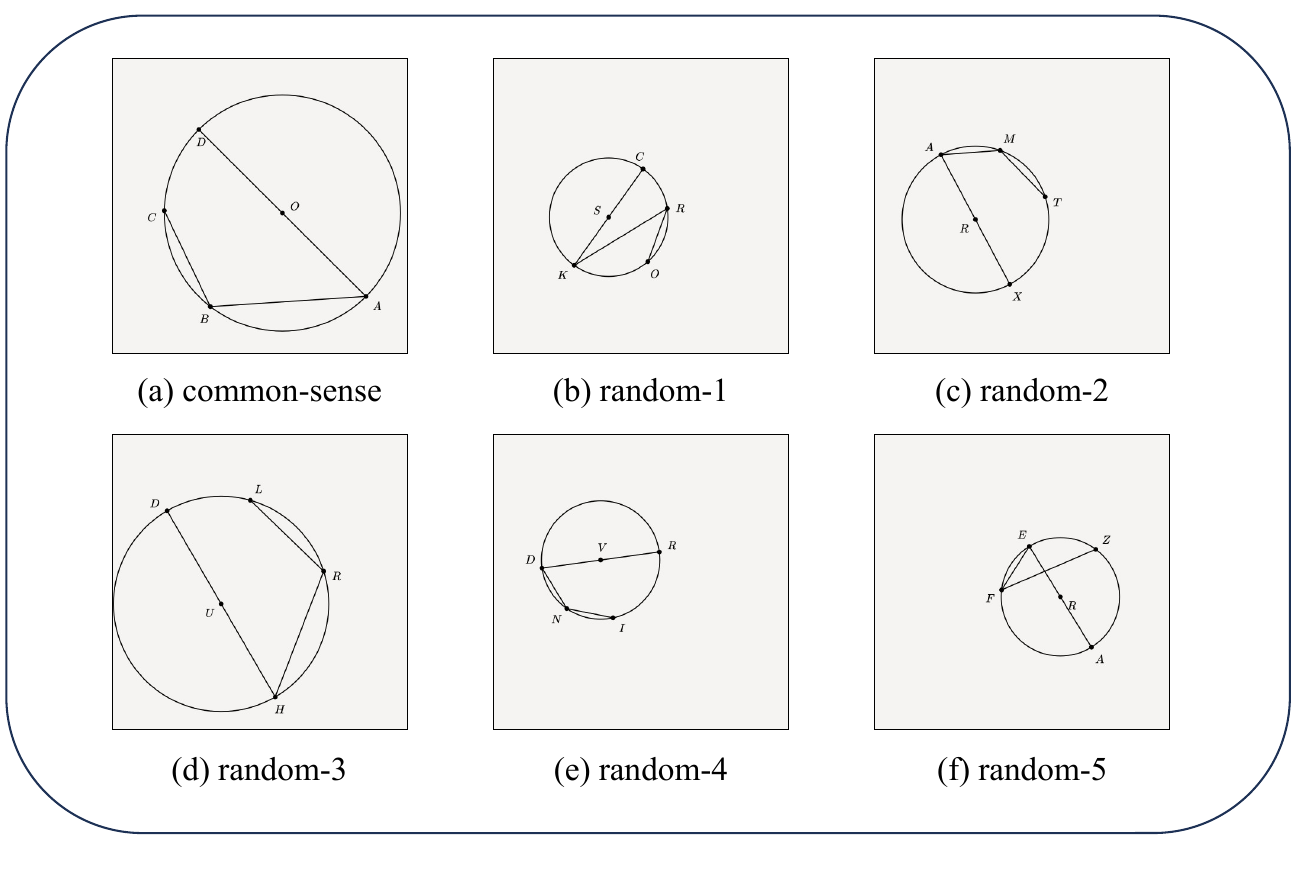}
   \caption{Different pictures generated by the same type of formal language. (a) is the common-sense  identifier of synthetic data, (b)-(f) are random  identifiers of synthetic data.}
   \label{fig:synthetic}
\end{figure*}

\textbf{Implementation Details.}
To ensure reproducibility, we set the random seed to 42.
During testing, we use a maximum output length of 1024 tokens, a batch size of 1, and a temperature of 0. To ensure output stability, we employ a 3-shot demonstration setting, as illustrated in~\Cref{fig:prompt-qa}. To systematically evaluate the impact of synthetic data on the geometric structure understanding of MLLMs, we fine-tuned the models with consistent hyperparameters on the training set and conducted quantitative evaluation on the GeoRef dataset. Both Supervised Fine-Tuning (SFT) and Reinforced Fine-Tuning (RFT) were trained for 500 steps.

\begin{figure*}[h]
  \centering\
   \includegraphics[width=1\linewidth, height=0.9\textheight, keepaspectratio]{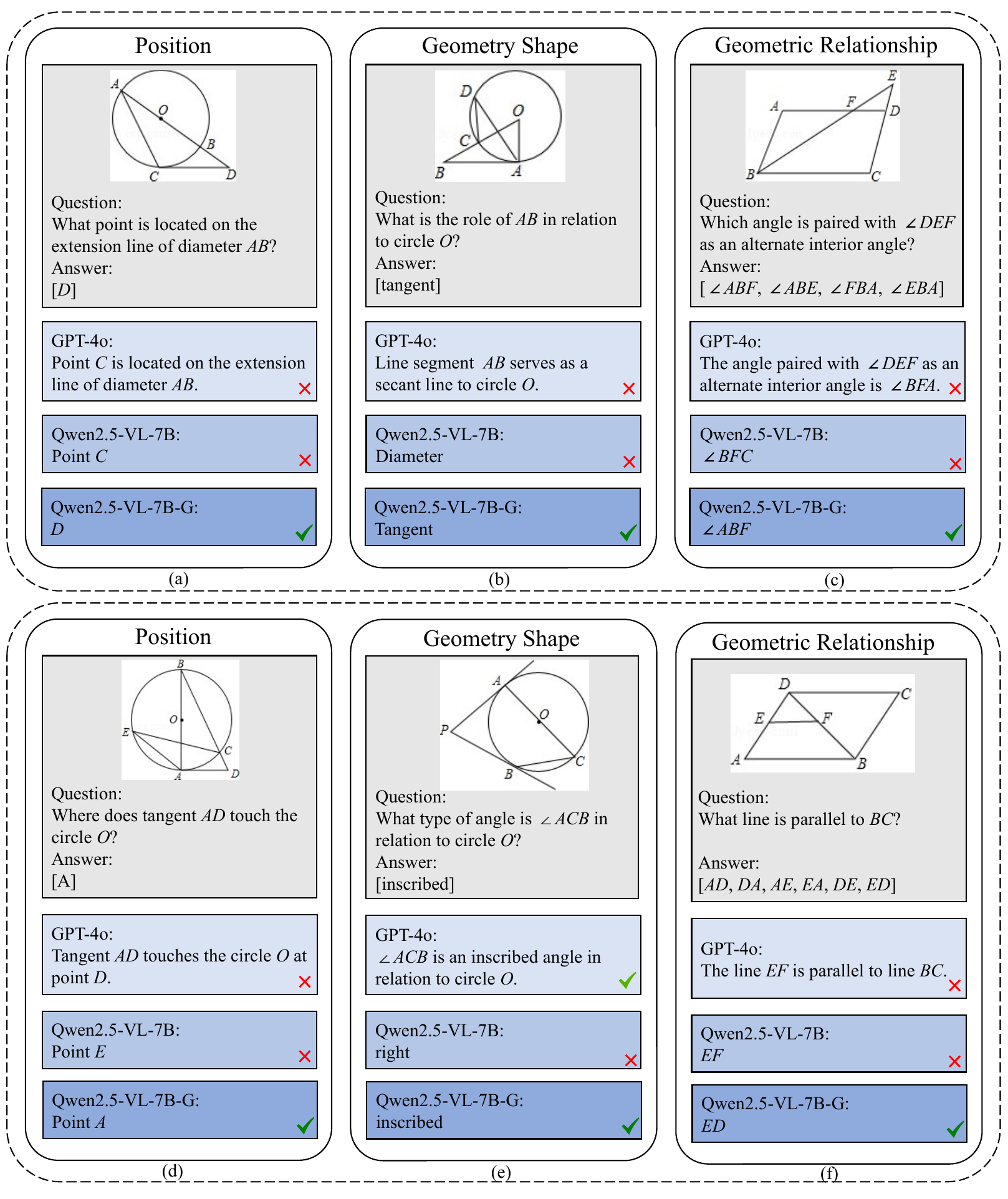}

   \caption{Quantitative assessment of the MLLMs on representative tasks from GeoRef. Qwen2.5-VL-7B-G denotes Qwen2.5-VL-7B fine-tuned on our synthesized training data.}
   \label{fig:fig6}
\end{figure*}

\begin{figure*}[h]
  \centering\
   \includegraphics[width=1\linewidth, height=1\textheight, keepaspectratio]{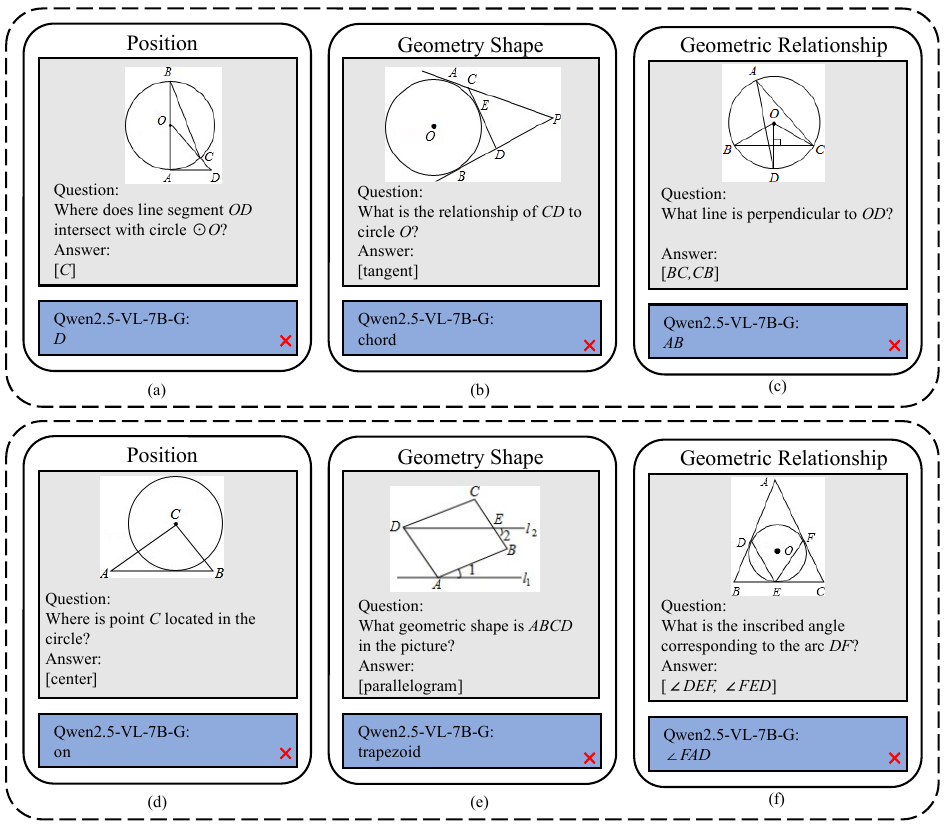}

   \caption{Failure cases of fine-tuning model Qwen2.5-VL-7B-G on GeoRef benchmark.}
   \label{fig:fig7}
\end{figure*}

We adopt LoRA~\cite{hu2022lora} fine-tuning for all training phases. During SFT, MiniCPM-V-2.6 and Qwen2.5-VL-7B utilize the LLaMA-Factory framework~\cite{zheng2024llamafactory}, while InternVL-2.5-8B is trained using its official repository. The fine-tuning parameters were set as follows: LoRA rank = 64 (lora target for all), per device batch size = 1, gradient accumulation steps = 2, learning rate = 1e-5, using a cosine learning rate scheduler with a warm-up ratio of 0.1. The entire fine-tuning process was conducted on 2 GPUs.

For RFT, we maintained the same experimental setup. Specifically, Qwen2.5-VL-7B adopts the VLM-R1 framework~\cite{shen2025vlm} and uses LoRA for GRPO training. The fine-tuning parameters were set as follows: LoRA rank = 64 (freeze vision modules), num generations = 8, batch size = 8, gradient accumulation steps = 2, learning rate = 1e-5, using a cosine learning rate scheduler with a warm-up ratio of 0.1, and training steps = 500. The entire fine-tuning process was conducted on 2 GPUs.

\section{Qualitative Visualization of the Geometric REC Results}\label{sec:case}
To validate the effectiveness of training data optimization in improving model performance, we compare our enhanced model, Qwen2.5-VL-7B-G, against the leading closed-source MLLM, GPT-4o, and the baseline model, Qwen2.5-VL-7B. As shown in~\Cref{fig:fig6}, geometric image test cases across different tasks demonstrate that the optimized model notably improves multimodal geometric spatial understanding.~\Cref{fig:fig6} (a) illustrates a simple point localization problem, where the task is to identify the point on the extended line. However, both GPT-4o and Qwen2.5-VL-7B mistakenly assume that point \(C\) lies on the extension of line \(AB\), while Qwen2.5-VL-7B-G provides the correct answer.~\Cref{fig:fig6} (d) involves the location of a geometrically significant point—the point of tangency. GPT-4o erroneously identifies point \(D\) as the tangency point, and Qwen2.5-VL-7B incorrectly designates point \(E\), which is not even on tangent \(AD\), as the tangency point on the circle. In contrast, Qwen2.5-VL-7B-G correctly determines that point \(A\) is the position where tangent \(AD\) touches circle \(O\). These two examples demonstrate that even for the most basic point localization tasks within geometric spatial understanding, the performance of general models is unsatisfactory and leaves considerable room for improvement.

In geometric shape analysis, GPT-4o misclassified segment \(AB\) as a secant of circle \(O\) in problem~\Cref{fig:fig6} (b), even though \(OA\) exhibited a clear perpendicularity to \(AB\). Meanwhile, Qwen2.5-VL-7B incorrectly identified \(AB\) as a diameter, despite the evident misalignment of point \(B\) with the circle. The fine-tuned Qwen2.5-VL-7B-G successfully corrected this issue, accurately determining that AB is a tangent to circle \(O\). 
In problem~\Cref{fig:fig6} (e), Qwen2.5-VL-7B also exhibited a logical error, mistakenly associating \(\angle ACB\) with a perpendicular relationship to circle \(O\), which is geometrically impossible. 
Notably, both GPT-4o and Qwen2.5-VL-7B-G correctly recognized the geometric properties of the angle, accurately identifying \(\angle ACB\) as an inscribed angle and correctly establishing its relationship with circle \(O\), demonstrating a precise understanding of the inscribed angle theorem.

When discussing geometric relationships, accurately understanding the connections between line segments and angles is crucial. Take the question in~\Cref{fig:fig6} (c), which asks about angle relationships. The correct answer should be \(\angle ABF\). GPT-4o provided an incorrect answer, while Qwen2.5-VL-7B even identified an angle that doesn’t exist in the diagram. In contrast, Qwen2.5-VL-7B-G correctly identified \(\angle ABF\) as the alternate interior angle of \(\angle DEF\), demonstrating its superiority in geometric reasoning.
Another example is from~\Cref{fig:fig6} (f), which asks about the parallel relationship between line segments. According to the properties of a parallelogram, opposite sides are parallel, so the correct answer should be \(AD\) or a segment of \(AD\). Neither GPT-4o nor Qwen2.5-VL-7B answered correctly, but Qwen2.5-VL-7B-G correctly identified \(ED\) (a segment of \(AD\)) as parallel to \(BC\).

The comparative results confirm the significant impact of training data optimization on enhancing MLLMs' geometric spatial understanding.

\textbf{Failure Cases of the Fine-tuned MLLM.}~\Cref{fig:fig7} presents empirical cases where the Qwen2.5-VL-7B-G model generated incorrect answers. Our study reveals that the model still exhibits errors caused by hallucinations. Moreover, its performance declines significantly when handling complex geometric images, highlighting the ongoing technical challenges faced by vision-language models in interpreting images with multiple spatial relationships and abstract features.

\clearpage
\newpage



\end{document}